\documentclass[10pt,twocolumn,letterpaper]{article}

\usepackage{wacv}
\usepackage{times}
\usepackage{epsfig}
\usepackage{graphicx}
\usepackage{amsmath}
\usepackage{amssymb}
\usepackage[dvipsnames]{xcolor}
\usepackage{multirow}
\usepackage{tabularx}

\newif\ifdraft\draftfalse

\ifdraft

\definecolor{greyblue}{rgb}{0.1,0.6,0.5}
\newcommand\Lars[1]{\textcolor{blue}{#1}}
\newcommand\yj[1]{\textcolor{ForestGreen}{#1}}
\newcommand\hj[1]{\textcolor{Sepia}{#1}}
\newcommand\nj[1]{\textcolor{red}{#1}}

\else

\newcommand\Lars[1]{#1}
\newcommand\yj[1]{#1}
\newcommand\hj[1]{#1}
\newcommand\nj[1]{#1}

\fi
  {\begin{list}{}%
          {\setlength{\leftmargin}{#1}}%
          \item[]%
  }
  {\end{list}}

\wacvfinalcopy 

\def\wacvPaperID{} 

\ifwacvfinal\fi
\begin{document}

\title{SINet: Extreme Lightweight Portrait Segmentation Networks with Spatial Squeeze Modules and Information Blocking Decoder }

\author{
Hyojin Park \\
Seoul National University\\
{\tt\small wolfrun@snu.ac.kr}
\and
Lars Lowe Sj{\"o}sund \\
 Clova AI, NAVER Corp \\
{\tt\small lars.sjosund@navercorp.com}
\and
YoungJoon Yoo \\
 Clova AI, NAVER Corp \\
{\tt\small youngjoon.yoo@navercorp.com}
\and
Nicolas Monet \\
NAVER LABS Europe\\
{\tt\small nicolas.monet@naverlabs.com}
\and
Jihwan Bang \\
 Search Solutions, Inc \\
{\tt\small jihwan.bang@navercorp.com}
\and
Nojun Kwak  \\
Seoul National University\\
{\tt\small nojunk@snu.ac.kr}
}
\maketitle
\ifwacvfinal\thispagestyle{empty}\fi

\begin{abstract}
Designing a lightweight and robust portrait segmentation algorithm is an important task for a wide range of face applications. 
However, the problem has been considered as a subset of the object segmentation problem and less \nj{handled} in the semantic segmentation field.
Obviously, portrait segmentation has its unique requirements.  
First, because the portrait segmentation is performed in the middle of a whole process of many real-world applications, it requires extremely lightweight models. 
Second, there has not been any public datasets in this domain that contain a sufficient number of images  with unbiased statistics.
To solve the \Lars{first} problem, we introduce \Lars{the} new extremely lightweight portrait segmentation model SINet, \Lars{containing an information blocking decoder and spatial squeeze modules. }
The information blocking decoder \nj{uses} \Lars{confidence estimates to recover local spatial information without spoiling global consistency. }
The spatial squeeze module \Lars{uses multiple receptive fields }
to cope with various size\Lars{s} of consistency in the image.
\Lars{To tackle the second problem, }we propose a simple method to create additional portrait segmentation data which can improve accuracy on the EG1800 dataset.
In our qualitative and quantitative analysis on the EG1800 dataset, we show that our method outperforms various existing lightweight segmentation models.
Our method reduces the number of parameters from $2.1 M$ to $86.9 K$ (around 95.9\% reduction), while maintaining the accuracy under \nj{an} 1\% margin from the state-of-the-art portrait segmentation method. 
We also show our model is successfully executed \Lars{on a} real mobile device with 100.6 FPS.
\hj{In addition, }
we demonstrate that our method 
\Lars{can be used} for general \nj{semantic} segmentation on the Cityscapes dataset.
The code and dataset are available in https://github.com/HYOJINPARK/ExtPortraitSeg .

\end{abstract}


\begin{figure}[t]
\begin{center}
\begin{tabular}{c}
    \includegraphics[width=0.95\linewidth]{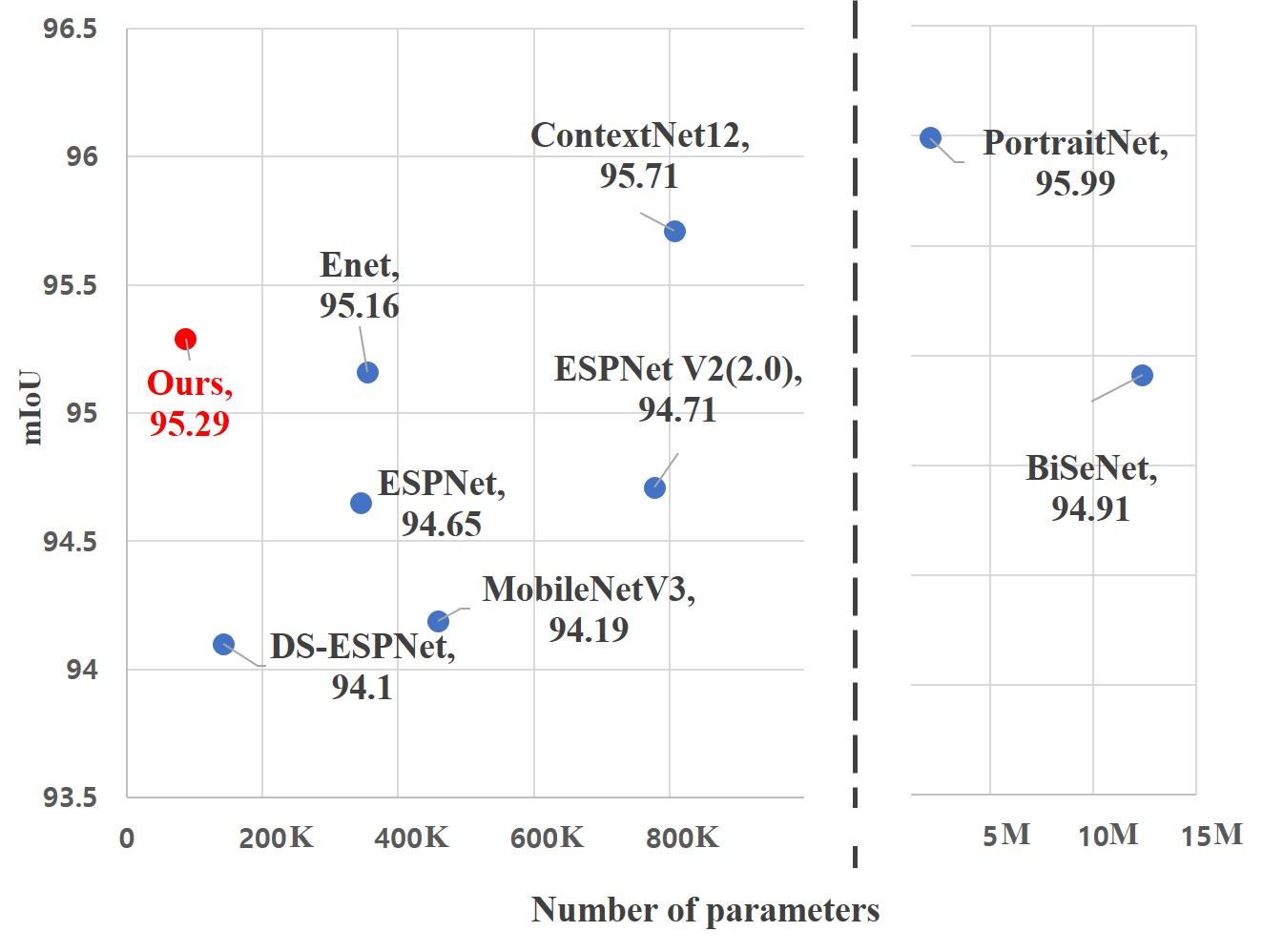}  \\ 
    
        \end{tabular}%
\end{center}
   \caption{Accuracy (mIoU) vs. complexity (number of parameters) on the EG1800 validation set. Our proposed SINet has high accuracy with small complexity.}
\label{fig:teaser2}
\end{figure}

\section{Introduction}
\label{sec:intro}

\yj{Developing} algorithms \yj{targeting} 
face data has been considered \nj{as} an important task in the computer vision field, and \yj{many related vision algorithms including} detection, recognition, \yj{and} key-point extraction are actively studied.
Among them, \Lars{portrait segmentation is commonly used} \nj{in real\hj{-world} applications such as}
background editing, security checks, and face resolution enhancement~\cite{shen2016automatic, zhang2019portraitnet}, 
\Lars{
giving rise to the need for fast and robust segmentation models}.

\begin{figure}[t]
\begin{center}
\begin{tabular}{c c}
    \includegraphics[width=0.3\linewidth]{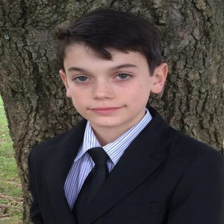} & \includegraphics[width=0.6\linewidth]{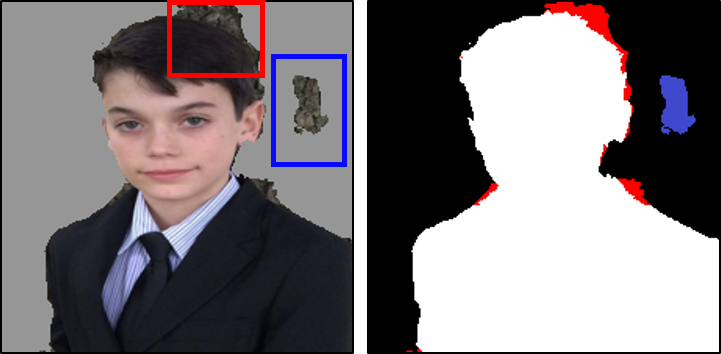} \\ 
    (a) Input image & (b) Typical segmentation errors\\ 
    \includegraphics[width=0.3\linewidth]{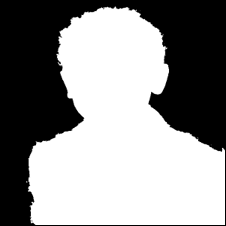} & \includegraphics[width=0.6\linewidth]{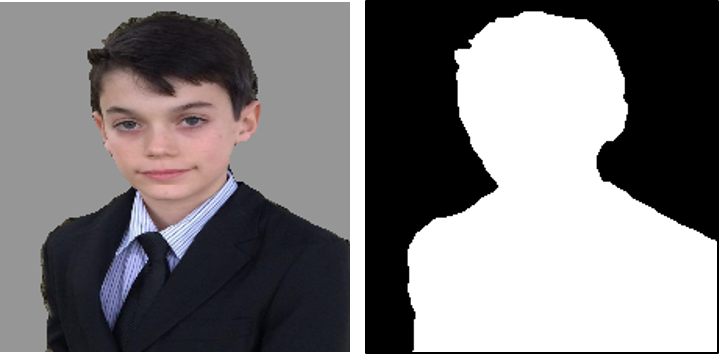} \\ 
    (c) Ground truth & (d) Example of Ours \\
        \end{tabular}%
\end{center}
   \caption{Typical \nj{e}xample\Lars{s} of \Lars{segmentation errors. }
   \nj{(b) T}he wood region is failed to \nj{be suppressed} and the boundary of \Lars{the} boy's hair is not sharp\nj{ly segmented}. \nj{(d)} Our method solve\nj{s} the problem without heavy computation.}
\label{fig:teaser1}
\end{figure}

The challeng\yj{ing} point \yj{of the} segmentation task is that the model have to solve two \nj{contradictory}
problem\Lars{s} \yj{simultaneously; }
(1) 
\Lars{Handling} long-range dependencies or global consistency and (2) preserving detailed local information.
Figure \ref{fig:teaser1} show\Lars{s two common segmentation errors.} 
First, \Lars{the} blue blob in Figure \ref{fig:teaser1} (b) \Lars{is classified as foreground, }
even \Lars{though} it is easily recognized as a wood region.
The reason for this problem \nj{is} that the segmentation model fails to get global context information \nj{which prevents} wrong representation.
Second, the red blob\Lars{s} in Figure \ref{fig:teaser1} (b) show \Lars{the model's failure to accurately segment fine details. }
The lateral part of hair needs fine segmentation due to \nj{its} small size and similar color \nj{to} the wood. The model is not able to produce a sharp segmentation image because of the lack of detail information about hair.
\hj{This is because the usage of stride convolution or pooling layer. These techniques induces the model can capture global information by enlarging receptive field size. However, the local information might be destroyed.}

\Lars{Researchers have developed several strategies to solve these problems}
\nj{and the first}
\Lars{one is to produce multiple receptive fields for each layer.} 
This multi-path structure is able to enhance both global and local information\Lars{, but comes at the cost of increased latency due to fragmented parallel operations \cite{ma2018shufflenet}. }
Another method is using \Lars{a} two-branch network, which consists of a deeper branch that is employed to \Lars{produce} global context, and a shallow branch 
\Lars{that preserves detailed local features by keeping high resolution} \cite{poudel2018contextnet, poudel2019fast, yu2018bisenet}.
Even though the shallow branch 
has few convolutional layers, it is computationally heavy due to its high-resolution feature maps.
Also, this method has to extract features two times, once for each branch.

\Lars{The portrait segmentation problem comes with a set of additional challenges.}
The first \Lars{one being the small amount of available data.} 
The EG1800 dataset~\cite{shen2016automatic}, an accessible public portrait segmentation dataset, contains only around 1,300 training images, and \Lars{has large biases with regard }
to attributes such as race, age, and gender. 
Second, portrait segmentation is usually used just as one of several steps in real-world applications. 
Since many of these applications run on mobile devices, the segmentation model needs to be lightweight to ensure real-time speeds.
Researchers have developed plenty of lightweight segmentation methods, but most of them are still not lightweight enough for portrait \nj{segmentation} tasks. \Lars{For example, \nj{PortraitNet~\cite{zhang2019portraitnet},} the current state-of-the-art model on the EG1800 dataset, has} 
$2.1 M$ parameters. A few examples \Lars{of} general lightweight segmentation model\Lars{s} are ESPNetV2~\cite{mehta2018espnetv2} with $0.78M$ parameters, and MobileNet V3~\cite{howard2019searching} with $0.47M$ parameters. 


 In this paper, we propose a new extremely lightweight portrait segmentation model \nj{called SINet} with \Lars{an} information blocking decoder structure and spatial squeeze module\Lars{s} (S2-module).
\Lars{Furthermore, we collect additional portrait segmentation data to overcome the aforementioned dataset problems. }
The proposed SINet has 86.9K parameters, \nj{achieving} 100.6 FPS in iPhone XS 
without any low-floating point operation\Lars{s or} pruning method\Lars{s}. 
\nj{Compar\Lars{ed} with \Lars{the} baseline model, PortraitNet, which has 2.1M parameters,} the accuracy degradation is just under $1\%$ on the EG1800 dataset, as can be seen in Figure~\ref{fig:teaser2}.

\noindent
\nj{Our contributions can be summarized as follows:}
 (1) 
\yj{We introduce the information blocking scheme to the decoder.}
\yj{It measures} 
\Lars{ the confidence in a low-resolution feature map, and blocks the influence of high-resolution feature maps in highly confident pixels. This prevents noisy information to ruin already certain areas, and allows the model to focus on regions with high uncertainty. }
We show that this information blocking decoder \Lars{is} robust to translation and can be applied to general segmentation tasks.
(2) We propose a spatial squeeze module (S2-module), an efficient multi-path network for feature extraction. 
\yj{Existing} multi-path structures \yj{deal with the} various size of long-range dependenc\Lars{ies} by managing \nj{multiple} receptive \nj{fields}.
However, \Lars{this} increases \nj{latency in real implementation}\Lars{s, due to having unsuitable structure with regard to kernel launching and synchronization. } 
To mitigate this problem, we squeeze the spatial resolution from each feature map by average pooling, \yj{and show that this is more effective} \nj{than adopting multi-receptive fields.} 
(3) The public portrait segmentation dataset has a \yj{small number of images} \nj{compared} to other  segmentation datasets, \yj{and \Lars{is} highly biased.}
We \yj{propose a} simple and effective data generation method to \yj{augment} the EG1800 dataset with \nj{a} \Lars{significant amount} of images.

\section{Related Work}
\label{sec:related}


\noindent
\textbf{Portrait Application: } 
PortraitFCN+~\cite{shen2016automatic} built a portrait dataset from Flickr and proposed a portrait segmentation model based on FCN~\cite{long2015fully}.
After that, PortraitNet proposed a \yj{real-time} portrait segmentation model with higher accuracy than PortraitFCN+. 
\cite{orrite2019portrait} \yj{integrated} two different segmentation \yj{schemes} \yj{from} Mask R-CNN and DensePose, and generated matting refinement based on FCN.
\cite{du2019boundary} \yj{introduced a }
boundary-sensitive kernel to enhance semantic boundary shape information.
\Lars{While these works achieved good segmentation results, their models are \nj{still} too heavy \nj{for} embedded systems.}

\noindent
\textbf{Global consistency:} 
Global consistency \Lars{and} long range of dependenc\Lars{ies} \yj{are} critical factors \yj{for} the segmentation task\Lars{, and models without a large enough receptive field will produce error\nj{-}prone segmentation maps. }
\Lars{One way of creating a large receptive field is to use large kernels. However, this is not suitable for lightweight models}
\yj{due to the\Lars{ir} large number of parameters}.
\Lars{Another method is to reduce \nj{the size of feature maps} through downsampling, but this leads to difficulties in segmenting} small or narrow object\yj{s}. 

To resolve this problem\yj{,} dilated convolution\Lars{s} (or atrous convolution\Lars{s})
\nj{have been introduced as} 
\Lars{an} effective solution to get \yj{a} large receptive field \Lars{while} 
preserving localization information \cite{yu2015multi, chen2018deeplab}, \nj{keeping the} same \yj{amount of} computation \Lars{as} the normal convolution.
However, \Lars{as the dilation rate is increased the count of valid weights decreases, to finally degenerate to a $1\times 1$ convolution~ \cite{chen2017rethinking}. }
Also, \Lars{the} grid effect degrade\yj{s the} segmentation result with checker\yj{-board} pattern.
Another method is \yj{to use} spatial pyramid pooling to get \Lars{a} large\yj{r} receptive field.
The spatial pyramid pooling uses different size\Lars{s} of pooling and concatenate\Lars{s} each \yj{resultant} feature map \yj{to obtain}  \nj{a} multi-\nj{scale} receptive field.
Similarly, \Lars{the} Atrous Spatial Pyramid Pooling layer~\cite{deeplabv3plus2018} replaces the pooling \Lars{with} dilated convolution\Lars{s} to get \Lars{a} \nj{multi-scale} representation. 
To get \Lars{a} multi-scale representation, some \nj{works use} \Lars{a} multi-path structure for feature extraction \cite{mehta2018espnet, mehta2018espnetv2, park2018concentrated}.
Each module split\yj{s} \Lars{the} input feature map and translate\yj{s the} feature map with \nj{a} differen\nj{t} dilat\Lars{ion} rate. 
This method is well 
\Lars{suited} for lightweight models, \yj{but suffers from high \Lars{latency. }}
Recently, 
\Lars{the} asymmetric non-local block~\cite{zhu2019asymmetric} \Lars{was} proposed, inspired by \Lars{the} non-local block~\cite{wang2018non} and spatial pyramid pooing.
\nj{Because the} non-local block calculate\yj{s} all \nj{the pairwise pixel dependencies, it is computationally heavy}.
Asymmetric non-local block approximate\nj{s} the calculation with spatial pyramid pooling. 
However, \yj{the computation\Lars{al} cost is still \Lars{too} large \nj{to fit a} 
lightweight model.}
Recently, some works adopt average pooling to reduce complexity more \cite{wang2019elastic, li2019hbonet}.

\noindent
\textbf{Detail local information:}
Recovering detail\Lars{ed} local information is crucial to generat\nj{ing} sharp segmentation map\Lars{s}.
\yj{Conventionally}, \nj{an e}ncoder-\nj{d}ecoder structure based on \nj{d}econvolution (or \nj{transposed convolution}) is applied~\cite{long2015fully, noh2015learning}.
\nj{By concatenating the} high-resolution feature, \nj{they} recover \nj{the} original resolution step by step.
Also, some works use global attention for upsampling. 
The feature pyramid attention \cite{li2018pyramid} use\nj{s} global pooling to enhance \nj{the} high resolution feature map from \nj{the} low-resolution.
However, the attention vector can not reflect \nj{the local information well} due to global pooling.
Recently, \nj{the} two-branch method is suggested \nj{for} better segmentation.
ContextNet~\cite{poudel2018contextnet} and FastSCNN~\cite{poudel2019fast} designed \nj{a two\yj{-path} network, each branch of which is for} global context and detailed information, respectively.
BiSeNet~\cite{yu2018bisenet} also proposed \yj{a similar} two-path network for preserving spatial information as well as acquiring a large enough receptive field.
However, it \nj{needs to} calculate feature\nj{s} \nj{twice}\Lars{, once} for each branch.



\begin{figure*}[t]
\begin{center}
\begin{tabular}{c}
    \includegraphics[width=0.90\linewidth]{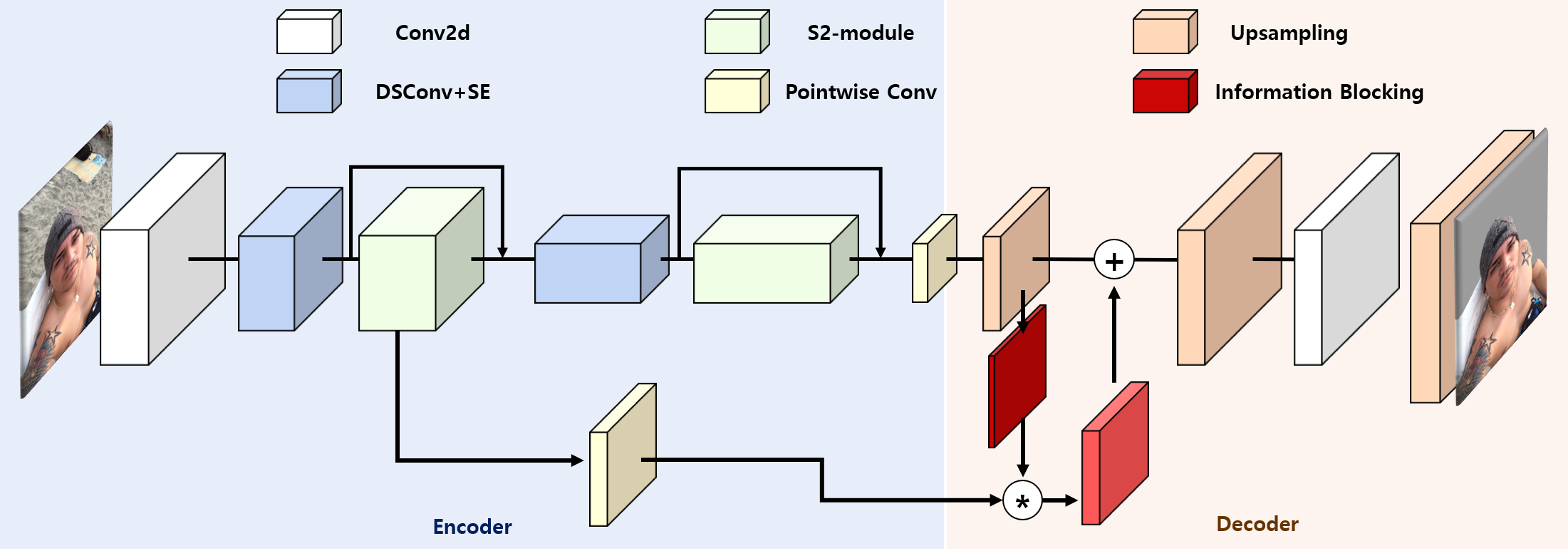} \\ 
    (a)  \\
        \end{tabular}%
\end{center}
   \caption{The overall architecture of SINet. The \nj{S2-}module is \nj{the} bottleneck in SINet and \nj{the} information blocking decoder makes fine segmentation results. DSConv+SE means depthwise separable convolution with a Squeeze-and-Excit\Lars{ation block}}
\label{fig:arch}
\end{figure*}

\section{Method}
\label{sec:method}

In this section, we explain the \yj{structure of the} proposed SINet 
\nj{which} consists of a Spatial squeeze module and a \nj{I}nformation blocking decoder. 
The spatial squeeze block (S2-module) \yj{handles} global consistency by \yj{using} \nj{the} multi-receptive field scheme, and squeeze\yj{s} the feature resolution to mitigate \Lars{the} high latency \Lars{of} multi-path structure\Lars{s}. 
The information blocking decoder \yj{is designed to} only take the necessary information from the high-resolution feature \Lars{maps} by 
\Lars{utilizing} the confidence \yj{score of} the low-resolution feature \Lars{maps}. 
The information blocking in the decoder is important for increasing robustness \yj{regarding} translation (Section \ref{IB_dnc}) and \nj{the} S2-module \yj{can} handle global consistency without heavy computation \nj{(Section \ref{S2})}. 
\nj{We also} demonstrate a simple data generation framework to solve the lack of data in two situations: 1) having human segmentation ground truths \nj{and} 2) having only raw images (Section \ref{sec:data}). 

\subsection{Information Blocking Decoder }
\label{IB_dnc}

\begin{figure}[t]
\begin{center}

\begin{tabular}{c}
    \includegraphics[width=0.95\linewidth]{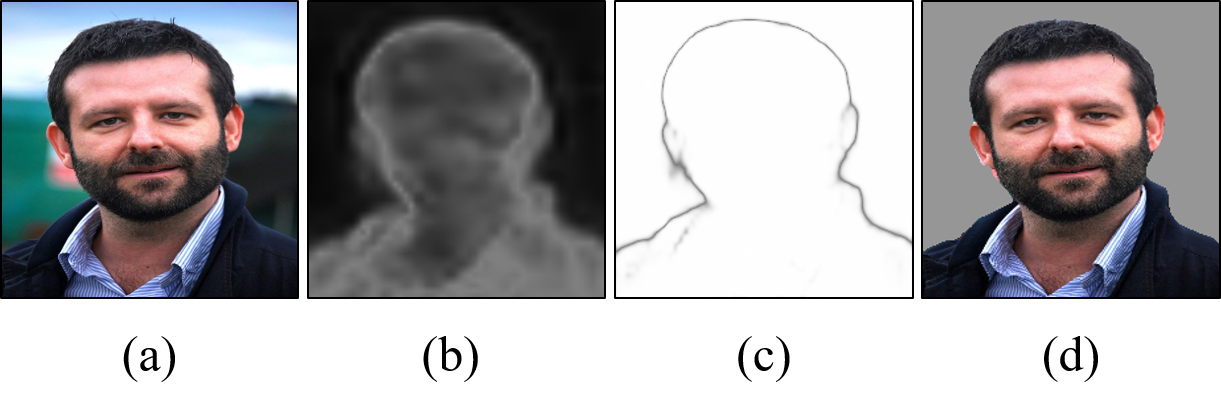}\\

        \end{tabular}%
\end{center}
\vspace{-4mm}
   \caption{From left to right,(a) an input image, (b) an information blocking map, (c) a confidence map of last block in model and (d) 
  \Lars{the segmentation results from the proposed method.}
   The information blocking map helps to prevent inflow of nuisance features\Lars{, and makes the model focus more on regions with high uncertainty.}}
\label{fig:confmap}
\end{figure}
An encoder-decoder structure is the \yj{most commonly used structure} for segmentation.
An encoder extracts semantic features \yj{of the incoming}  images according to semantic information, and a decoder recovers \yj{detailed} local information and resolution of the feature map. 
\yj{For designing the decoder, bilinear upsampling or \nj{transposed convolution} upsampling blocks are commonly used to expand the low-resolution feature map\Lars{s} from the encoder.}
\yj{Also, recent works \cite{mehta2018espnet, park2018concentrated, howard2019searching, deeplabv3plus2018}} re-use additional high-resolution feature maps from the encoder to make more accurate segmentation results.
\nj{To t}he best of our knowledge, \yj{most studies} take all the information of high-resolution feature maps from the encoder by conducting concatenation, element-wise summation, or by enhancing high-resolution feature maps via attention vector\Lars{s} from low-resolution.
However, using \yj{the} high-resolution feature maps means that we give nuisance local information, which is already removed \nj{by the encoder.} 
Therefore, we have to take only the necessary clue and avert the nuisance noise.

Here, we introduce a new concept of \Lars{a} decoder structure \yj{using} information blocking.
We measure the confidence score in the low-resolution feature map and block the information flow \Lars{from} the high-resolution feature into the region \nj{where the encoder successfully segmented with high confidence.} 
The information blocking process removes nuisance information from the image and make\nj{s} the high-resolution feature map \nj{concentrate only on} the low confidence region\Lars{s}.

Figure \ref{fig:arch} shows the \nj{overall architecture of SINet and the} detailed process of the information blocking decoder. 
\nj{T}he model projects the last \Lars{set of} feature map\Lars{s} \nj{of the encoder to the size of} the number of \nj{classes} by a pointwise convolution and uses a bilinear upsampling to make the same resolution \Lars{as} \nj{the} high-resolution \nj{target} \Lars{segmentation map. }
The model employs a softmax function to get a probability of each class and calculates each pixel's confidence score \nj{$c$} by taking maximum value \nj{among} the probabilities of each class.
Finally, we generate an information blocking map by \nj{computing} $(1-c)$. 
\Lars{We perform pointwise multiplication between the information blocking map and the high-resolution feature maps. This \nj{ensure\Lars{s}}} that low confidence regions get more information from the high-resolution feature map\Lars{s}, \Lars{while} high confidence regions keep the\Lars{ir} original value\Lars{s} \nj{in the subsequent pointwise addition operation}.

Figure \ref{fig:confmap} (b) is an \yj{example of the} information blocking map, and (c) \nj{is} the confidence map from the \Lars{model output. }
As shown in Figure \ref{fig:confmap} \nj{(b)}, the boundary \nj{and clothing} ha\Lars{ve} high uncertainty 
\Lars{while} the inner part\nj{s} of the foreground and background already ha\nj{ve} a high confidence score.
\nj{This indicates that} the high uncertainty regions need more detailed local information to \nj{reduce} uncertainty.
However, the inner part\nj{s} of the face, such as \Lars{the} beard and nose, \nj{do} not need to get more local information for making a segmentation map.
If the local information \Lars{was} embedded, it \yj{\nj{could} be harmful} to \Lars{the} 
global consistency due to nuisance information as noise.
\nj{In the final confidence map of the model (Figure. \ref{fig:confmap} (c)), 
the} uncertainty region of the boundary \Lars{has} shrunk, and the confidence score of the inner part is \nj{highly} improved.




\subsection{Spatial Squeeze module}
\label{S2}

\begin{figure*}[t]
\begin{center}
\begin{tabular}{c}
    \includegraphics[width=0.90\linewidth]{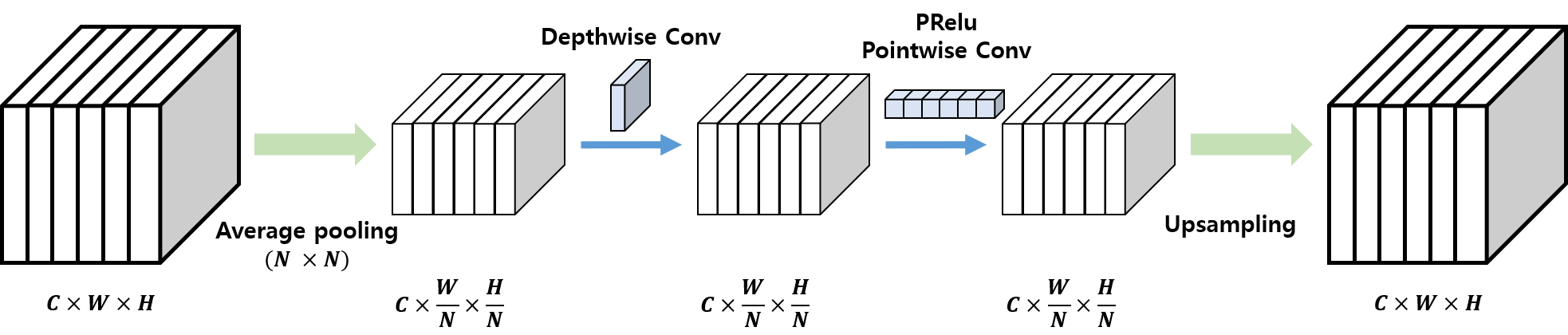} \\ 
    (a) Spatial Squeeze Block (S2-block).     \\

        \includegraphics[width=0.90\linewidth]{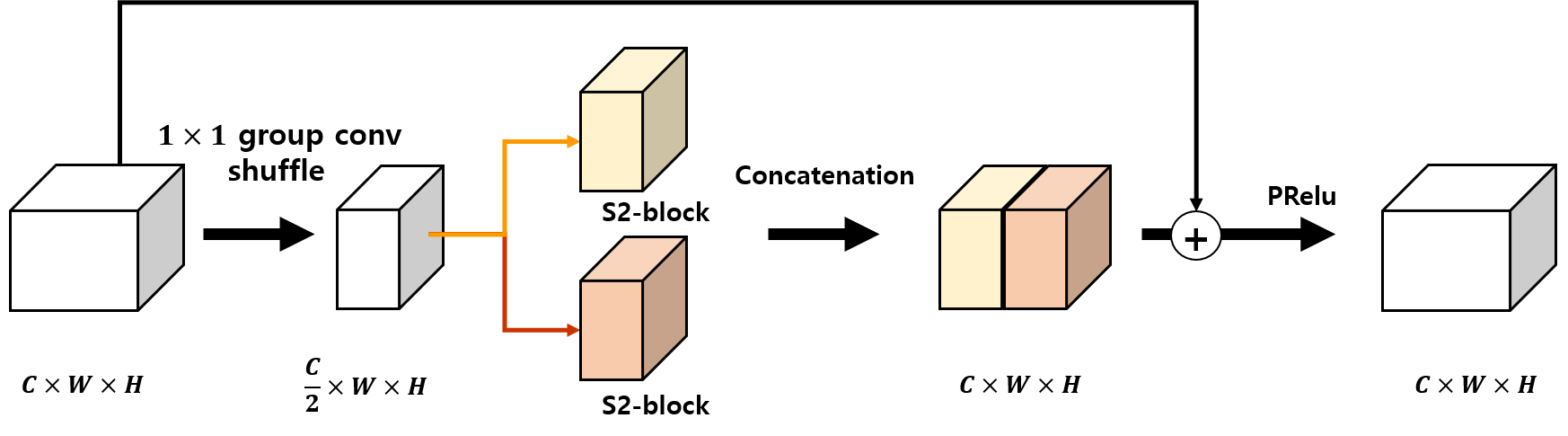} \\ 
    (b) Spatial Squeeze Module (S2-module).  \\
        \end{tabular}%
\end{center}
   \caption{(a) An input feature map is squeezed by \Lars{an} $N \times N$ average pooling before a depthwise convolution. Then, a bilinear upsampling recovers \Lars{the original input} resolution of \Lars{the} feature map\Lars{s}. (b) \Lars{The} S2-module has multi-receptive structures by \Lars{using} different combination\Lars{s} of convolution kernel\Lars{s} and pooling for \Lars{the} S2-block\Lars{s}.}
\label{fig:s2}
\vspace{-2mm}
\end{figure*}

\begin{table}[t]
\begin{center}

    \begin{tabular}{l | c c c c}
     \hline
     Dilated rate & rate=2 & rate=6 & rate=12 & rate=18 \\
     \hline \hline
     Latency (ms) & 6.7  & 6.63 & 11.84 & 12.03 \\
     \hline
    \end{tabular}%
    \end{center}
      \caption{\Lars{Latency running depthwise separable dilated convolution with different dilation rates on an iPhone XS.} 
      The input size is $128\times120\times120$.  \Lars{Additional experiments with other input sizes are reported in the supplementary material.}}
  \label{tab:dilation_time}%
\end{table}%

\nj{A m}ulti-path structure have an advantage of high accuracy with less parameters~\cite{wang2019elastic, szegedy2017inception, szegedy2017inception, xie2017aggregated}, but \nj{it} suffers from \nj{increased} latency \nj{proportional to} the number of sub-paths \cite{ma2018shufflenet}.
The proposed spatial squeeze module \textit{(S2-module)} resolves this problem and \nj{Figure} \ref{fig:s2} shows the structure.
We utilize average pooling for adjusting \nj{the} size of \nj{the} receptive field and \nj{reducing the latency}.

The S2-module is also \Lars{following a} kind of split-transform-merge scheme like \cite{mehta2018espnet, mehta2018espnetv2, park2018concentrated} for covering multi-receptive field with two spatial squeeze blocks \textit{(S2-block)}.
First, we uses a pointwise convolution \Lars{to reduce the number of feature maps by half.}
For further reduction of computation, we use a group pointwise convolution with channel shuffle.
The reduced feature map\Lars{s} pass through each S2-block, \Lars{and the results are merged through concatenation.}
We also adopt a residual connection between \Lars{the} input feature map and the merged feature map.
Finally, PRelu is utilized for non-linearity.

For S2-block, we select average pooling rather than \nj{dilated} convolution for making a multi-receptive field structure for two reasons.
First, the latency time is affected from the dilated rate, as shown in Table \ref{tab:dilation_time}, and dilated convolution can not be free from \nj{the problem of grid effects} \cite{park2018concentrated, wang2017understanding}.
Second, the multi-path structure is not friendly to GPU parallel computing~\cite{ma2018shufflenet}. 
Thus, we squeeze the resolution of each feature map to avoid the sacrifice of the latency time. 
\Lars{The} S2-block squeezes the resolution of a feature map by an average pooling\Lars{, with kernel size up to 4}.
\Lars{Then, a depthwise separable convolution with \nj{the} kernel size $3$ or $5$ is used. }
Between the depthwise convolution and \Lars{the} pointwise convolution, we use a PRelu non-linear activation function.
Empirically, 
\Lars{placing} the pointwise convolution before or after the bilinear upsampling does not have a critical effect on the accuracy.
\Lars{Therefore, we put it }
before the bilinear upsampling \Lars{to further reduce computation. }
We also insert a batch normalization layer after the depthwise convolution and the bilinear upsampling.

\subsection{Network Design for SINet}
\label{SINet}

\begin{table}[t]
\tiny
  \begin{center}
      \begin{tabular}{c|c|c|c|c}
          & Input & Operation & Output &  \\
          \hline 
    1     &  $3\times224\times224$ & CBR    & $12\times112\times112$   & Down sampling \\
    2     &  $12\times112\times112$ & DSConv+SE &  $16\times56\times56$ & Down sampling \\
    3     &  $16\times56\times56$ & SB module &  $48\times56\times56$ & [k=3, p=1], [k=5, p=1] \\
    4     & $48\times56\times56$& SB module & $48\times56\times56$ & [k=3, p=1], [k=3, p=1] \\
    5     & $64\times56\times56$ & DSConv+SE & $48\times28\times28$  & Concat [2, 4], Down sampling \\
    6     &  $48\times28\times28$ & SB module &  $96\times28\times28$ & [k=3, p=1], [k=5, p=1] \\
    7     & $96\times28\times28$ & SB module & $96\times28\times28$ & [k=3, p=1], [k=3, p=1] \\
    8     & $96\times28\times28$ & SB module & $96\times28\times28$ & [k=5, p=1], [k=3, p=2] \\
    9     & $96\times28\times28$ & SB module & $96\times28\times28$ & [k=5, p=2], [k=3, p=4] \\
    10    &$96\times28\times28$ & SB module & $96\times28\times28$ & [k=3, p=1], [k=3, p=1] \\
    11    & $96\times28\times28$ & SB module & $96\times28\times28$ & [k=5, p=1], [k=5, p=1] \\
    12    & $96\times28\times28$ & SB module & $96\times28\times28$ & [k=3, p=2], [k=3, p=4] \\
    13    & $96\times28\times28$& SB module & $96\times28\times28$ & [k=3, p=1], [k=5, p=2] \\
    14    & $144\times28\times28$& 1x1 conv & $\#class\times28\times28$ & Concat [5, 13]\\
   
    \end{tabular}%
    \end{center}
    \caption{Detailed setting\Lars{s} for \Lars{the} SINet encoder. k denotes \Lars{the} kernel size of \Lars{the} depthwise convolution and p \Lars{denotes the} kernel size of average pooling \Lars{the} S2-block. }
  \label{tab:setting}%
  \vspace{-3mm}
\end{table}%

In this \nj{part}, we explain the overall structure of SINet.
SINet uses S2-module\Lars{s} \Lars{as} bottleneck\nj{s} and depthwise separable convolution \textit{(ds-conv)} \Lars{with stride $2$} for reducing \nj{the resolution of} \Lars{feature maps}.
Empirically, applying the S2-module with stride $2$ \Lars{for} downsampling improves accuracy, but we found that \Lars{it} 
has \nj{longer} latency time than S2-module with stride $1$  under the same output size conditions.
Therefore, \Lars{for downsampling}, \nj{instead of the S2-module with stride 2, we use} ds-conv with Squeeze-and-Excite block\Lars{s}.
\Lars{For the first bottleneck we use two sequential S2-modules and for the second bottleneck we use eight. }
The detailed setting of the S2-module is described in Table \ref{tab:setting}.
\Lars{We add a residual connection for each bottleneck, concatenating the bottleneck input with its output.}
\nj{A} $3\times3$ convolution \nj{is used} for classification and \Lars{finally} bilinear upsampling \nj{is applied} \Lars{to recover the original input resolution. }

We found that a weighted auxiliary loss for the boundary part is helpful in improving \Lars{the} accuracy.
The final loss is as \nj{follows:}
\begin{equation}
\begin{aligned}
& \mathcal{B} = (f \oplus y^*) - (f \ominus y^*) \\
& Loss = CE_{i \in \mathcal{P}}(y_i^*, \hat{y}) + \lambda CE_{j \in \mathcal{B}}(y_j^*, \hat{y}_j).
\end{aligned}
\label{eq:loss}
\end{equation}
\nj{Here,} $f$ is a $15\times15$ filter used for the \nj{morphological} dilation \nj{($\oplus$)} and erosion \nj{($\ominus$)} operations.
$\mathcal{P}$ denotes \Lars{all} \nj{the} pixels of \Lars{the} ground truth, and $\mathcal{B}$ \Lars{denotes the} pixels in \Lars{the} boundary area 
as defined by the morphology operation. 
$y^*$ is a binary ground truth value and $\hat{y}$ is a predicted label from a segmentation model. \nj{$\lambda$ is a hyperparameter that controls the balance between the loss terms.}

\begin{table*}[t]
\small
  \begin{center}
\begin{tabular*}{0.99\textwidth}{@{\extracolsep{\fill}}| l |  cccccc|}
\hline
    Method & Parameters (M) & FPS & FLOPs (G)   & F1-score& mIOU  & \nj{mIOU}~\cite{zhang2019portraitnet} \\
    \hline
    \hline
    Enet (2016) \cite{paszke2016enet} & 0.355 & 8.06     & 0.346 &    0.917   & 95.16 & 96 \\
    BiSeNet (2018) \cite{yu2018bisenet} & 0.124 &  2.99     & 2.31  &  0.908   & 94.91 & 95.25 \\
    PortraitNet (2019)\cite{zhang2019portraitnet} & 2.08 & 3.30      & 0.325 & \textcolor{blue}{ 0.919 }    & \textcolor{blue}{95.99} & 96.62 \\
    ESPNet (2018)\cite{mehta2018espnet} & 0.345 &  6.99     & 0.328 &   0.883    & 94.65 & - \\
    DS-ESPNet  & 0.143  &    7.75      & 0.199 & 0.866    & 94.10  & - \\
    DS-ESPNet(0.5)  & \textcolor{blue}{0.064} &  9.26     & 0.139 &  0.859     & 93.58 & - \\
    ESPNetV2(2.0) (2019)\cite{mehta2018espnetv2}  & 0.778 &  3.65     & 0.231 &   0.872    & 94.71  & - \\
     ESPNetV2(1.5) (2019)\cite{mehta2018espnetv2} & 0.458  &   4.95    & 0.137 &    0.861   & 94.00 & - \\
    ContextNet12 (2018)\cite{poudel2018contextnet} & 0.838  &  1.55     & 1.87  &   0.896    & 95.71 & - \\

    MobileNetV3 (2019)\cite{howard2019searching} &   0.458     & 10.87      & 0.066 &  0.854     & 94.19  & - \\
    \textbf{SINet(Ours)}  & \textbf{0.087}  &   \textcolor{blue}{\textbf{12.35 }}   & \textcolor{blue}{\textbf{0.064}} &  \textbf{0.884 }    & \textbf{94.81}  & - \\
    \textbf{SINet+(Ours)}  & \textbf{0.087} &  \textcolor{blue}{ \textbf{12.35  }}  &\textcolor{blue}{ \textbf{0.064}} &  \textbf{0.892 }    & \textbf{95.29}  &  -\\
    
    \hline
    \end{tabular*}%
    \end{center}
     \caption{EG1800 validation results for the proposed SINet and other segmentation models. DS denotes depth-wise separable convolution. We measure FPS \Lars{on an} Intel Core 15-7200 CPU environment with \Lars{input size} $224 \times 224$. \nj{The results in the last column are from} \Lars{the} PortraitNet~\cite{zhang2019portraitnet} paper. SINet+ is \nj{the} result of using \Lars{the} augmented dataset \nj{as descibed in} Section \ref{sec:data}}
  \label{tab:exp}%
\end{table*}%

\subsection{Data Generation}
\label{sec:data}
Annotating data often comes with high costs, and the annotation time per instance varies a lot depending on the task type. For example, the annotation time per instance for PASCAL VOC \Lars{is estimated} to be 20.0 seconds for image classification and 239.7 seconds for segmentation, an order of magnitude difference as mentioned in \cite{bearman2016s} .
To mitigate the cost of annotation for portrait segmentation, we consider a couple of plausible situations: 1) having images with ground truth human segmentation.
2) having only raw images.
We make use of either an elaborate face detector model (case 1) or a segmentation model (case 2) for generating pseudo ground truths to each situation. 

When we have human images and ground truths, \yj{the only thing  we need is} a bounding box around the portrait area.
We took images from Baidu dataset~\cite{wu2014early}, which contains 
5,382 human full body segmentation images covering various poses, fashions and backgrounds.  
To get the bounding box and portrait area, we detect the face location of the images using a face detector~\cite{yoo2019extd}.
Since the face detector tightly bounds the face region, we increase the bounding box size to include parts of the upper body and background before cropping the image and ground truth segmentation.

We also create a second \yj{augmentation} from portrait images scraped from the web, applying a more heavyweight segmentation model to generate pseudo ground truth segmentation masks.
This segmentation model consists of a  DeepLabv3+~\cite{deeplabv3plus2018} architecture with a SE-ResNeXt-50~\cite{xie2017aggregated} backbone. The model is pre-trained on ImageNet and finetuned on a proprietary dataset containing around 2,500 fine grained human segmentation images. The model is trained for general human segmentation rather than for the specific purpose of portrait segmentation. 

Finally, human annotators just check the quality of each pseudo ground truth image, removing obvious failure cases. 
This method reduces the annotation effort per instance from several minutes to 1.25 seconds by transforming the segmentation task into a binary classification task.

\section{Experiment}
\label{sec:exp}

\begin{figure*}[t]
\begin{center}
\begin{tabular}{c}
    \includegraphics[width=0.90\linewidth]{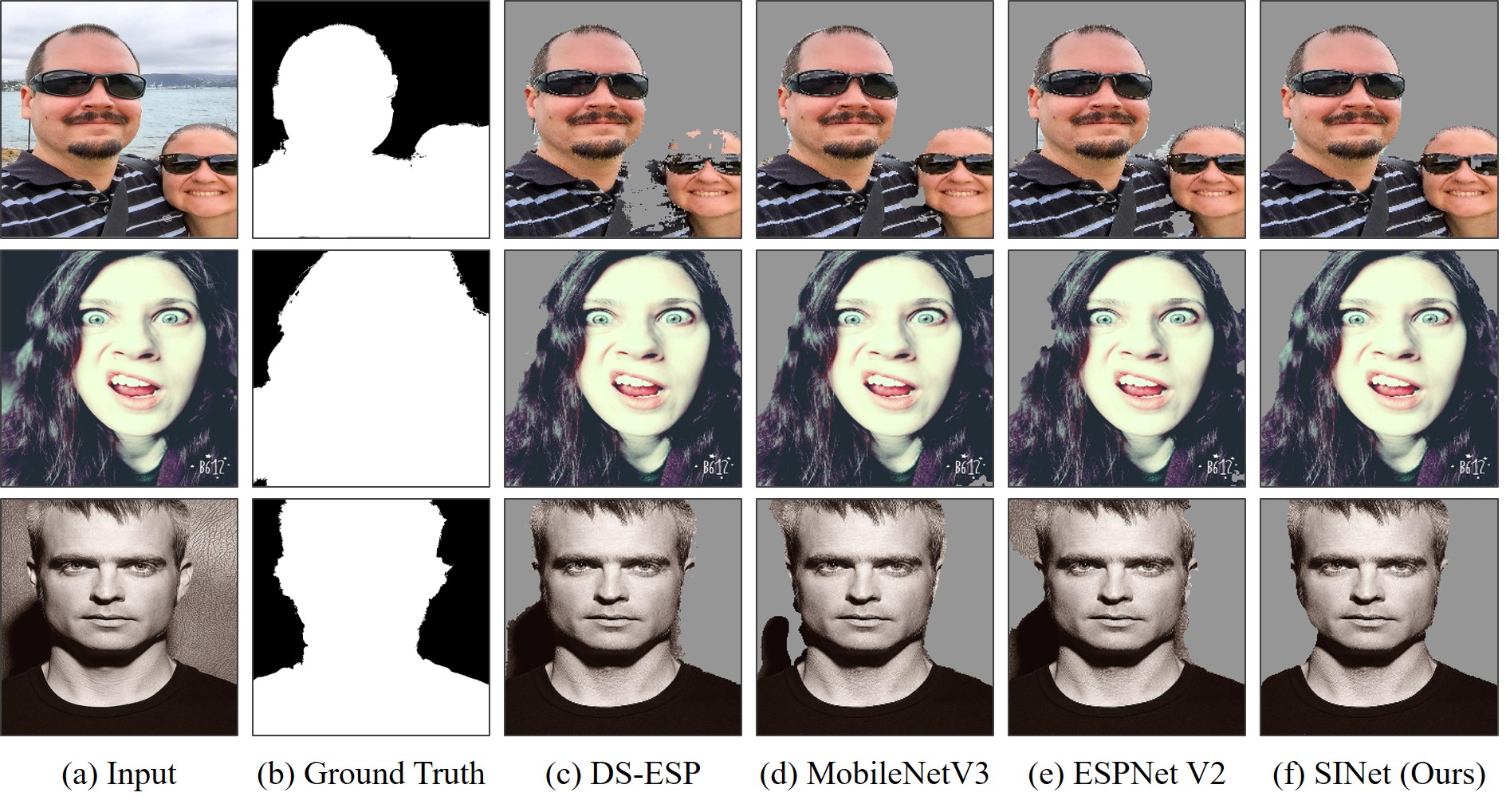}  \\ 
        \end{tabular}%
\end{center}
\vspace{-4mm}
   \caption{Qualitative comparison results on the EG1800 validation dataset.}
\label{fig:ex}
\end{figure*}

We evaluated the proposed method on the public dataset EG1800~\cite{shen2016automatic}, which collected images from Flickr with manually annotated labels.
The dataset has a total of $1,800$ images and is divided into $1,500$ train and $300$ validation images.
However, we could access only 1,309 images for train and 270 for validation, since some of the URLs are broken. 
We built \Lars{an} additional 10,448 images \Lars{using} the proposed data generation method mentioned in Section ~\ref{sec:data}.

We trained our model using ADAM optimizer with initial learning rate to $7.5\mathrm{e}{-3}$, \Lars{and} weight decay to $2\mathrm{e}{-4}$, \Lars{for a} total \Lars{of} 600 epochs.
We followed \Lars{the} data augmentation method \Lars{in} \cite{zhang2019portraitnet} with $224 \times 224$ image\Lars{s}.
We used a two-stage training method;
 for the first 300 epochs, we only trained until \Lars{the} encoder with the batch size \Lars{set} to $36$.
Then, \Lars{we initialize the encoder with the best parameters }
from the previous step, \Lars{and} trained the overall SINet model for an additional 300 epochs with the batch size to $24$.
We evaluated our model followed by various ablations using mean intersection over union (mIoU) and F1-score in \Lars{the} boundary part, \Lars{and} compared with SOTA portrait segmentation model\Lars{s} including other lightweight segmentation models.
\Lars{To define the boundary region, we subtract the eroded ground truths from the dilated ground truths, using a kernel with size $15\times15$. } 
We \Lars{demonstrated} the robustness of \Lars{the} information blocking decoder \Lars{on} randomly rotated EG1800 validation images, and \Lars{the} importance of multi-receptive structure \Lars{on} EG1800 validation images in Section \ref{sec:ablation}
Also, we show\Lars{ed} that the proposed method can be used \Lars{for} general task\Lars{s} by \Lars{evaluating it on the} Cityscape\Lars{s} dataset.

\subsection{Evaluation Results on the EG1800 Dataset}
\label{sec:EG1800}
We compared the proposed model to PortraitNet\cite{zhang2019portraitnet}, which has SOTA accuracy in the portrait segmentation field. 
Since some sample URLs in the EG1800 dataset are missing, we re-trained the PortraitNet follow\Lars{ing} the original method in paper and \Lars{using the} official code \Lars{on} the remaining samples in EG1800 dataset.
PortraitNet compared their work to BiseNet and Enet. Therefore, we also re-trained BiSeNet and ENet following the method of PortraitNet for a fair comparison.
As shown in Table \ref{tab:exp}, the accurac\Lars{ies} of the re-trained \Lars{models} 
are slightly decreased due to the reduced size of the training dataset.
We measure\yj{d} latency time on an Intel Core i5-7200U CPU environment with the PyTorch framework on \Lars{an} LG gram laptop.

Among the compar\Lars{ed} methods, DS-ESPNet has the same structure \Lars{as} ESPNet, with only changing the standard dilated convolutions of the model into depth-wise separable dilated convolutions.
For ESPNetV2~(2.0) and ESPNetV2~(1.5), we changed the \Lars{number of} channel\Lars{s} of \Lars{the} convolution\Lars{al layers} to reduce the model size as following official code. 
We also reduced the \Lars{number of channels for the convolutions} 
in \Lars{the} DS-ESPNet~(0.5) by half from the original model to make it 
less than 0.1M parameters and 0.2G FLOPs.
The original ContextNet used 4 pyramid poolings but we used only 3 
due to \yj{the} small \Lars{feature map} size.

From Table~\ref{tab:exp}, we see that our proposed method \yj{achieved} comparable or better performance than the other models, \Lars{while having} less parameters \Lars{and} FLOPs\yj{,} and \Lars{higher} FPS.
The SOTA PortraitNet showed the highest accuracy in all the experimental results, and has achieved even better performance than the heavier BiSeNet.
However, PortraitNet requires a large number of parameters, which is a disadvantage for using it on smaller devices.
The proposed SINet has reduced the number of parameters by 95\%, and FLOPs by 80\% compared to PortraitNet, while maintaining accuracy.
ESPNet and ESPNet V2 have similar accuracy, but showed a trade-off between the number of parameters and FLOPs.
ESPNet V2 has more parameters than ESPNet, but ESPNet needs more FLOPs than ESPNet V2.
Enet shows better performance than both models but requires more FLOPs.
In \Lars{our} comparison, \yj{the} proposed method has less number of parameters and FLOPs, but \Lars{still} \yj{achieved} better accuracy than ESPNet and ESPNet V2.
In particular, our SINet has the highest accuracy in an extremely lightweight environment.
Figure \ref{fig:ex} shows that the quality of our model is superior to other extremely lightweight models.

We compared the execution speed of the proposed model with SOTA segmentation model MobileNet V3 on an iPhone XS using \Lars{the} CoreML framework.
MobileNet V3 has 60.7 FPS, \Lars{and} \textbf{our SINet \Lars{has} 100.6 FPS}.
The FLOPs \Lars{in} MobileNet V3 and SINet are similar, but SINet is much faster than MobileNet V3.
We \yj{conjecture} that the SE block and h-swish activation function are the main reasons \Lars{for the} increase \Lars{in} latency in MobileNet V3.
In summary, the proposed SINet show\yj{ed} outstanding performance among the various segmentation model in terms of accuracy and speed.

\subsection{Ablation Study}
\label{sec:ablation}

\begin{figure}[t]
\begin{center}

\begin{tabular}{c}
    \includegraphics[width=0.95\linewidth]{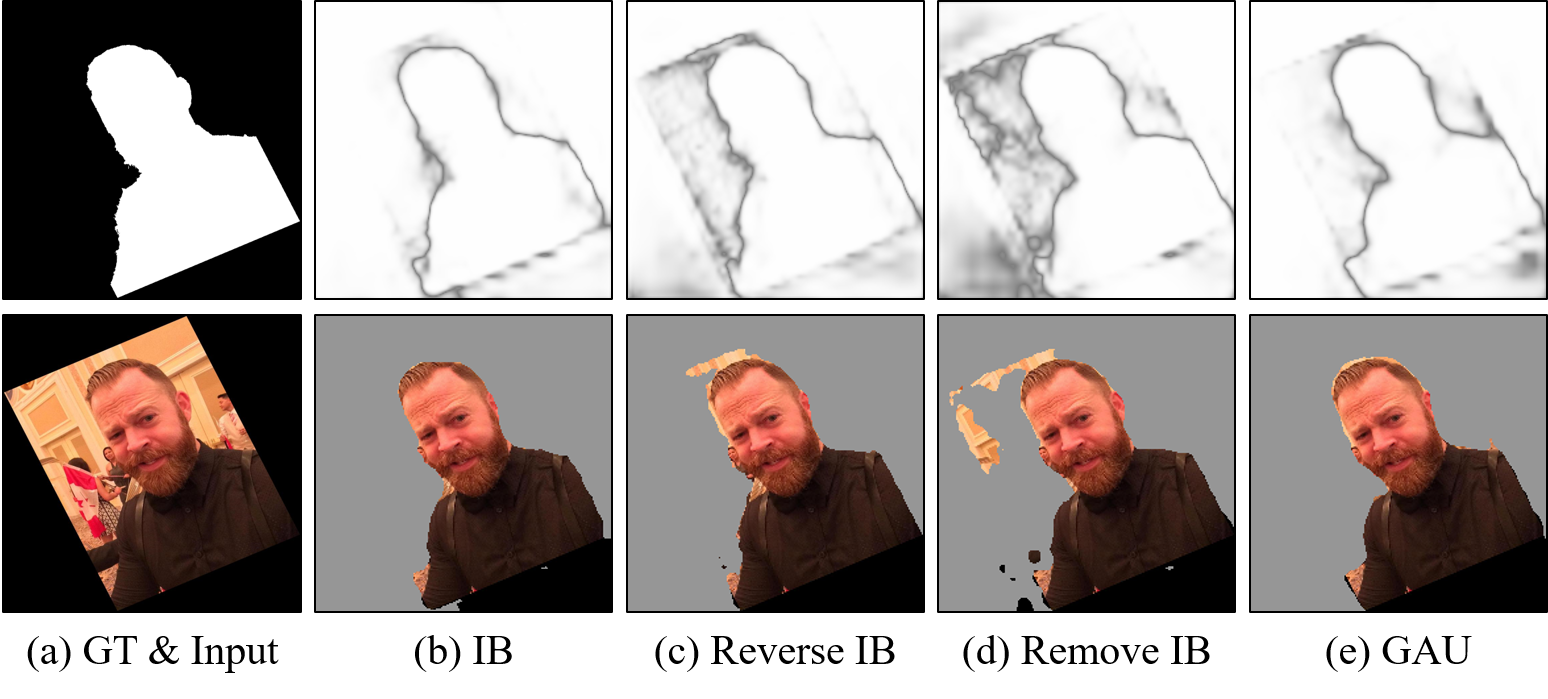}\\

        \end{tabular}%
\end{center}
\vspace{-2mm}
   \caption{Confidence maps of \nj{the} last features from \nj{the} model and segmentation results according to \nj{the decoding} method.}
\label{fig:ablation_IB}
\end{figure}

\begin{table}[t]
  \begin{center}
   \begin{tabular}{l | ccccc}
    \hline
          & Org    & $25^\circ$    & $45^\circ$   & $75^\circ$    & $90^\circ$ \\
          \hline \hline
    IB    & 94.81 & 88.62 & 84.74 & 79.05 & 76.85 \\
    Reverse IB & 94.68 & 84.06 & 80.58 & 73.88 & 73.09 \\
    Remove IB   & 94.74 & 85.11 & 81.53 & 75.36 & 72.99 \\
    GAU \cite{li2018pyramid}   & 94.77 & 86.33 & 82.80  & 75.85 & 73.96 \\

    \hline
    \end{tabular}%
 
  \end{center}
   
  \caption{mIOU results \Lars{when using} random rotation\Lars{s} ($-x^\circ \sim x^\circ $). Org denotes original validation image without any translation. IB denotes using information blocking map and Reverse IB enhances high-confidence regions rather than low-confidence ones.}
  \label{tab:IBablation}%
\end{table}%

\noindent
\textbf{Information blocking decoder: } 
Table \ref{tab:IBablation} show\yj{s} the accuracy improvement from \Lars{using the} information blocking decoder.
We randomly rotated validation images and evaluated mIOU \Lars{over the whole image. }
Reverse IB denotes that we \Lars{multiply the high-resolution feature maps with the confidence score $c$ instead of $(1 - c)$, thus enhancing high-confident pixels rather than low-confidence ones. }
Remove IB means that we \Lars{did not use any information blocking, and instead} conducted element-wise summation between \Lars{the} low-resolution feature map\Lars{s} and \Lars{the} high-resolution feature map\Lars{s} from a middle layer of a encoder.
GAU uses global pooling to enhance high-resolution feature map from low-resolution feature map before applying element-wise summation.
GAU has better performance than Reverse IB and Remove IB, but it still fails to get a tight boundary and to get better performances in translated images than IB.
\yj{From the result, we can see that} the information blocking decoder shows outstanding performance \Lars{compared to the} other methods.
\yj{Qualitatively}, it prevents segmentation \Lars{errors} of \Lars{the} background region as shown in Figure \ref{fig:ablation_IB}.

\noindent
\textbf{Multi-receptive field: } 
\begin{table}[t]
  \begin{center}
    \begin{tabular}{ cc| cc}
    \hline
    Convolution & Pooling &  mIOU & f1-score \\
    \hline \hline
    
    3     & 1     & 93.40 & 0.874 \\
    3     & 2     & 93.34 & 0.851 \\
    3     & 4     & 90.34 & 0.796 \\
    5     & 1     & 94.56 & 0.881 \\
    5     & 2     & 93.11 & 0.857 \\
    5     & 4     & 90.04 & 0.799 \\
    \hline
    \multicolumn{2}{c|}{SINet (Ours)} & 94.81  & 0.884 \\

    \end{tabular}%
    \end{center}
    
    \caption{Ablation study \yj{for} \Lars{the} S2-module. Our multi-receptive structure \yj{achieved} better accuracy than \Lars{the other settings. }}
     \label{tab:multiR}%
  \end{table}%
Table \ref{tab:multiR} shows \Lars{the} performance \yj{depending on the multi-receptive structures.}
SINet use\yj{d} various \Lars{combinations of} kernel size\Lars{s for} convolution and pooling.
We re-designed the S2-module \Lars{to always use the same kernel sizes within the S2-block for all convolutional and pooling layers respectively. }
As shown in Table \ref{tab:multiR}, our SINet \Lars{achieved} high\Lars{er} mIOU and F1-score than \yj{the} other combinations.
Therefore, a multi-receptive field structure has an advantage for accuracy than a single-receptive field one.

\subsection{General Segmentation Dataset}
\begin{table}[t]
\footnotesize
   \begin{center}
     
    \begin{tabular}{l | cccc}
    \hline
          & Param(M) & FLOP(f) & FLOP(h) & mIOU \\
          \hline\hline
    ESPNet & 0.36  & -     & 4.5   & 61.4 \\
    ContextNet14 & 0.85  & 5.63  & -     & 66.1 \\
    ESPNetV2 & 0.79  & -     & 2.7   & 66.2 \\
    MobileNetV2(0.35) & 0.16  & 2.54  & -     & 66.83(val) \\
    MobileNetV3-small & 0.47  & 2.9   & -     & 69.4 \\
    \textbf{SINet (Ours)}  & \textbf{ 0.12}  &  \textbf{1.2}   & -     &  \textbf{66.5} \\
    \hline
    \end{tabular}%
      \end{center}
    \caption{Semantic segmentation results on \Lars{the} Cityscapes test set. FLOP(f) means that \Lars{the number of} FLOPs \yj{was} measured with full-resolution input, $2048\times1024$. FLOP(h) denotes that \Lars{the number of} FLOPs \yj{was} measured with half-resolution, $1024\times512$.}
  \label{tab:addlabel}%
\end{table}%

\begin{table*}[t]
\footnotesize
  \begin{center}
     \begin{tabularx}{0.99\textwidth}{
      >{\centering\arraybackslash}m{5mm} 
     | >{\centering\arraybackslash}X 
     | >{\centering\arraybackslash}X 
     | >{\centering\arraybackslash}X 
     | >{\centering\arraybackslash}X }
   
          & Input & Operation & Output &  \\
          \hline \hline
    
    1     & $3\times1024\times2048$ & CBR   & $16\times512\times1024$ & Down sampling \\
    2     & $16\times512\times1024$ & DSConv + SE & $\#class\times256\times512$ & Down sampling \\
    3     & $\#class\times256\times512$ & DSConv + SE & $24\times128\times256$ & Down sampling \\
    4     & $24\times128\times256$ & SB module & $60\times128\times256$ & [k=3, p=1], [k=5, p=1] \\
    5     & $60\times128\times256$ & SB module & $60\times128\times256$ & [k=3, p=0], [k=3, p=1] \\
    6     & $60\times128\times256$ & SB module & $60\times128\times256$ & [k=3, p=0], [k=3, p=1] \\
    7     & $84\times128\times256$ & DSConv + SE & $60\times64\times128$ & Concat [3, 6] , Down sampling \\
    8     & $60\times64\times128$ & SB module & $84\times64\times128$ & [k=3, p=1], [k=5, p=1] \\
    9     & $84\times64\times128$ & SB module & $84\times64\times128$ & [k=3, p=0], [k=3, p=1] \\
    10    & $84\times64\times128$ & SB module & $84\times64\times128$ & [k=5, p=1], [k=5, p=4] \\
    11    & $84\times64\times128$ & SB module & $84\times64\times128$ & [k=3, p=2], [k=5, p=8] \\
    12    & $84\times64\times128$ & SB module & $108\times64\times128$ & [k=3, p=1], [k=5, p=1] \\
    13    & $108\times64\times128$ & SB module & $108\times64\times128$ & [k=3, p=1], [k=5, p=1] \\
    14    & $108\times64\times128$ & SB module & $108\times64\times128$ & [k=3, p=0], [k=3, p=1] \\
    15    & $108\times64\times128$ & SB module & $108\times64\times128$ & [k=5, p=1], [k=5, p=8] \\
    16    & $108\times64\times128$ & SB module & $108\times64\times128$ & [k=3, p=2], [k=5, p=4] \\
    17    & $108\times64\times128$ & SB module & $108\times64\times128$ & [k=3, p=0], [k=5, p=2] \\
    18    & $168\times64\times128$ & 1x1 conv & $\#class\times64\times128$ & Concat [7, 17] \\
    \end{tabularx}%
    \end{center}
    \caption{Detailed setting\Lars{s} for \Lars{the} SINet encoder. k denotes \Lars{the} kernel size of \Lars{the} depthwise convolution and p \Lars{denotes the} kernel size of average pooling \Lars{the} S2-block. }
  \label{tab:citysetting}%
\end{table*}%

We also demonstrate that our proposed method is suitable not only for \Lars{the} binary segmentation problem but also \yj{for} general segmentation problem\Lars{s} \yj{by testing the model} \Lars{on the} Cityscape\Lars{s} dataset.
We increased \Lars{the} number of layers and channels a little bit to cope with \yj{the increased} complexity  \yj{compared} to \Lars{the} binary segmentation task, and we factorized the depthwise convolution in \Lars{the} S2-block\Lars{s} for reducing \Lars{the} number of parameters.
\Lars{Here}, SINet has only 0.12M parameters and 1.2GFLOPs \Lars{for input of size} $2048\times512$, but our model \yj{showed} better accuracy than any other lightweight segmentaiton model except MobileNet V3 and MobileNet V2.
The accuracy of SINet decrease\Lars{s} by 2.9\% \Lars{with} respect to MobileNet V3, but the number of parameters and FLOPs are much lower than MobileNet V3. 
Table~\ref{tab:citysetting} is a detailed setting of the encoder model for the Cityscape segmentation. 

\label{city}
\section{Conclusion}
In this paper, we proposed \yj{an extremely lightweight portrait segmentation model,} SINet, which consists of \Lars{an} information blocking decoder and spatial squeeze module\Lars{s}.
SINet executes well in mobile device with 100.6FPS and has high accuracy with 95.29.
The information blocking decoder prevents nuisance information from high-resolution feature\Lars{s} and induce the model to concentrate more \Lars{on} high uncertainty region\Lars{s}.
The spatial squeeze module has multi-receptive field to handle \Lars{the} various sizes of global consistency in \Lars{an} image.
We \Lars{also} proposed a simple data generation framework covering the two situations: 1) having human segmentation ground truths 2) having only raw images. 
Not only on the specific portrait dataset but also on the general segmentation dataset, our model obtained outstanding performance compared to the existing lightweight segmentation models from the experiments.
The proposed method shows appropriate accuracy (66.5 \%) with only 0.12M number of parameter and 1.2G FLOP on the Cityscapes dataset. 

\label{sec:endLOL}

{\small
\bibliographystyle{ieee}
\bibliography{egbib}
}

\newpage

\section*{Appendix}

\subsection*{Additional Experiment Results}
\label{latency}

\begin{table}[ht]
\tiny
  \begin{center}

   \begin{tabular}{l| cccc | cccc}
    \hline
    Input Resolution & \multicolumn{8}{c}{$48\times48$} \\
      \hline
    Input Channel & \multicolumn{4}{c|}{32}        & \multicolumn{4}{c}{128} \\
          & Min  &  Mean & Max   & Total & Min  &  Mean & Max   & Total \\
          \hline \hline
    $d=2$ & 1.4   & 1.6   & 1.99  & 159.91 & 1.64  & 1.92  & 2.75  & 192.29 \\
    $d=6$  & 1.41  & 1.66  & 1.98  & 166.08 & 1.67  & 1.92  & 2.69  & 192.75 \\
    $d=12$  & 2.63  & 5.22  & 9.16  & 524.23 & 4.32  & 8.63  & 11.78 & 866.7 \\
    $d=18$  & 3.05  & 5.42  & 9.96  & 544.49 & 3.52  & 8.38  & 12.04 & 841.26 \\
    \hline
    Input Resolution & \multicolumn{8}{c}{$120\times120$} \\
      \hline
    Input Channel & \multicolumn{4}{c|}{32}        & \multicolumn{4}{c}{128} \\
         & Min  &  Mean & Max   & Total & Min  &  Mean & Max   & Total \\
           \hline \hline
    $d=2$  & 5.22  & 5.54  & 8.44  & 555.17 & 6.36  & 6.7   & 11.21 & 671.03 \\
    $d=6$  & 4.86  & 5.36  & 7.79  & 536.96 & 6.42  & 6.63  & 11    & 663.61 \\
    $d=12$  & 8.99  & 14.03 & 25.65 & 1403.78 & 9.2   & 11.84 & 15.18 & 1184.8 \\
    $d=18$  & 9.09  & 13.84 & 24.51 & 1384.74 & 9.48  & 12.03 & 15.45 & 1204.13 \\
      \hline
    Input Resolution & \multicolumn{8}{c}{$320\times320$} \\
      \hline
    Input Channel & \multicolumn{4}{c|}{32}        & \multicolumn{4}{c}{128} \\
          & Min  &  Mean & Max   & Total & Min  &  Mean & Max   & Total\\
           \hline \hline
    $d=2$  & 26.92 & 27.76 & 49.53 & 2777.01 & 36.33 & 37.23 & 73.88 & 3724.02 \\
    $d=6$  & 27.02 & 28.13 & 48.41 & 2813.39 & 36.25 & 37.48 & 73.78 & 3748.36 \\
    $d=12$  & 30.7  & 31.84 & 52.32 & 3184.4 & 46.31 & 47.74 & 78.9  & 4774.73 \\
    $d=18$  & 31.45 & 32.66 & 55.9  & 3267.09 & 46.54 & 47.84 & 79.86 & 4785.13 \\
    \hline
    \end{tabular}%
      \end{center}
 
  \caption{Latency time experiments about depthwise separable dilated convolution on iPhone CPU/GPU/NPU environment. All performance value are reported in milliseconds. \textit{\textbf{d}} denotes a dilated rate of convolution filter.}
   \label{tab:latency1}%
\end{table}%

\begin{table}[ht]
   \begin{center}
\scriptsize
       
    \begin{tabular*}{0.45\textwidth}{@{\extracolsep{\fill}}l ||  cccc}
          & \multicolumn{4}{c}{CPU} \\
         
    Model  &  Min  &  Mean & Max   & Total \\
    \hline  \hline 
    SINet & 9.69  & 9.8   & 12.1  & 980.11 \\
    MobileNet V3 & 15.81 & 16.07 & 19.2  & 1607.54 \\
    ESPNet V2 & 95.79 & 96.48 & 103.44 & 9649.16 \\
     \hline \hline
          & \multicolumn{4}{c}{CPU/GPU} \\
           Model  &  Min  &  Mean & Max   & Total \\
           \hline   \hline 
    SINet & 44.51 & 83.01 & 123.84 & 8303.74 \\
    MobileNet V3 & 48.9  & 110.53 & 145.87 & 11056.51 \\
    ESPNet V2 & 59.83 & 91.53 & 640   & 9154.22 \\
    \hline \hline
          & \multicolumn{4}{c}{CPU/GPU/NPU} \\
           Model  &  Min  &  Mean & Max   & Total \\
           \hline   \hline 
    SINet & 9.63  & 9.94  & 16.94 & 994.45 \\
    MobileNet V3 & 15.88 & 16.18 & 26.28 & 1619.12 \\
    ESPNet V2 & 36.87 & 37.65 & 46.6  & 3766.04 \\
    
    \end{tabular*}%
     \end{center}
  
    \caption{Latency time experiments under different environments. 
    We converted ESPNet V2 Pascal VOC segmentation model by CoreML framework from their official converting code for baseline of latency time in a mobile environment.  
    The input size of SINet and MobileNet V3 is $224 \times 224$ and the input size of ESPNet V2 is $256 \times 256$.
    The number of FLOPs of SINet, MobileNet V3 and ESPNet V2  are $0.064G$, $0.066$ and  $0.338G$.
    Our SINet is faster than MobileNet V3 and ESPNet V2 in all the environments.
    All performance value are reported in milliseconds}
    
    \label{tab:latency2}%
\end{table}%

The goal of this experiment is to show exact performances on different types of layers or models that run on iOS.
We are running on all the experiments with iPhone XS Max and iOS 13.1.2.
We did 100 iterations and waited 2 seconds between each run, and all performance values are reported in milliseconds. 
Each model in Table \ref{tab:latency1} are tested on CPU/GPU/NPU configuration and each model in Table \ref{tab:latency2} are executed in different configuration (CPU, CPU/GPU, CPU/GPU/NPU ).
As shown in Table \ref{tab:latency1}, a large dilated rate convolution layer is slower than others regardless of input size.
Table \ref{tab:latency2} proves that our SINet is faster than any other model in every configuration.
This result demonstrates that our model can be applied to any configurations well.

\end{document}